# The Anchors Hierarchy: Using the Triangle Inequality to Survive High Dimensional Data

**Andrew W. Moore**
Carnegie Mellon University
Pittsburgh, PA 15213
www.andrew-moore.net

## Abstract

This paper is about metric data structures in high-dimensional or non-Euclidean space that permit cached sufficient statistics accelerations of learning algorithms.

It has recently been shown that for less than about 10 dimensions, decorating *kd*-trees with additional "cached sufficient statistics" such as first and second moments and contingency tables can provide satisfying acceleration for a very wide range of statistical learning tasks such as kernel regression, locally weighted regression, k-means clustering, mixture modeling and Bayes Net learning.

In this paper, we begin by defining the *anchors hierarchy*—a fast data structure and algorithm for localizing data based only on a triangle-inequality-obeying distance metric. We show how this, in its own right, gives a fast and effective clustering of data. But more importantly we show how it can produce a well-balanced structure similar to a Ball-Tree (Omohundro, 1991) or a kind of metric tree (Uhlmann, 1991; Ciaccia, Patella, & Zezula, 1997) in a way that is neither "top-down" nor "bottom-up" but instead "middle-out". We then show how this structure, decorated with cached sufficient statistics, allows a wide variety of statistical learning algorithms to be accelerated even in thousands of dimensions.

## 1 Cached Sufficient Statistics

This paper is not about new ways of learning from data, but instead how to allow a wide variety of current learning methods, case-based tools, and statistics methods to scale up to large datasets in a computationally tractable fashion. A cached sufficient statistics representation is a data structure that summarizes statistical information from a large dataset. For example, human users, or statistical programs, often need to query some quantity (such as a mean or variance) about some subset of the attributes (such as size, position and shape) over some subset of the records. When this happens, we want the cached sufficient statistic representation to intercept the request and, instead of answering it slowly by database accesses over huge numbers of records, answer it immediately.

For all-discrete (categorical) datasets, cached sufficient statistics structures include frequent sets (Agrawal, Mannila, Srikant, Toivonen, & Verkamo, 1996) (which accelerate certain counting queries given very sparse high dimensional data), datacubes (Harinarayan, Rajaraman, & Ullman, 1996) (which accelerate counting given dense data up to about 7 dimensions), and all-dimensions-trees (Moore & Lee, 1998; Anderson & Moore, 1998) (which accelerate dense data up to about 100 dimensions). The acceleration of counting means that entropies, mutual information and correlation coefficients of subsets of attributes can be computed quickly, and (Moore & Lee, 1998) show how this can mean 1000-fold speed-ups for Bayes net learning and dense association rule learning.

But what about real-valued data? By means of *mrkd*-trees (multiresolution k-dimensional trees) (Deng & Moore, 1995; Moore, Schneider, & Deng, 1997; Pelleg & Moore, 1999, 2000) (an extension of *kd*-trees (Friedman, Bentley, & Finkel, 1977)) we can perform clustering, and a very wide class of non-parametric statistical techniques on enormous data sources hundreds of times faster than previous algorithms (Moore, 1999), but only up to about 8-10 dimensions.

This paper replaces the *kd*-trees with a certain kind of Metric tree (Uhlmann, 1991) and investigates the extent to which this replacement allows acceleration on real-valued queries in higher dimensions. To achieve this, in Section 3 we introduce a new tree-free ag-glomerative method for very quickly computing a spatial hierarchy (called the *anchors hierarchy*)—this is needed to help generate an efficient structure for the metric tree. And while investigating the effectiveness of metric trees we introduce three new cached sufficient statistics algorithms as examples of a much wider range of possibilities for exploiting these structures while implementing statistical and learning operations on large data.

## 2 Metric Trees

Metric Trees, closely related to Ball trees (Omohundro, 1991), are a hierarchical structure for representing a set of points. They only make the assumption that the distance function between points is a metric:

$$\forall x, y, z \; D(x, z) \leq D(x, y) + D(y, z)$$
$$\forall x, y \; D(x, y) = D(y, x)$$
$$\forall x \; D(x, x) = 0$$

Metric trees do not need to assume, for example, that the points are in a Euclidean space in which vector components of the datapoints can be addressed directly. Each node $n$ of the tree represents a set of datapoints, and contains two fields: $n_{\text{pivot}}$ and $n_{\text{radius}}$. The tree is constructed to ensure that node $n$ has

$$n_{\text{radius}} = \max_{x \in n} D(n_{\text{pivot}}, x) \qquad (1)$$

meaning that if $x$ is owned by node $n$, then

$$D(n_{\text{pivot}}, x) \leq n_{\text{radius}} \qquad (2)$$

If a node contains no more than some threshold $R_{\min}$ number of points, it is a leaf node, and contains a list of pointers to all the points that it owns.

If a node contains more than $R_{\min}$ points, it has two child nodes, called $Child_1(n)$ and $Child_2(n)$. The points owned by the two children partition the points owned by $n$:

$$x \in Child_1(n) \iff x \notin Child_2(n)$$
$$x \in n \iff x \in Child_1(n) \text{ or } x \in Child_2(n)$$

How are the child pivots chosen? There are numerous schemes in the Metric tree literature. One simple method is as follows: Let $f_1$ be the datapoint in $n$ with greatest distance from $n_{\text{pivot}}$. Let $f_2$ be the datapoint in $n$ with greatest distance from $f_1$. Give the set of points closest to $f_1$ to $Child_1(n)$, the set of points closest of $f_2$ to $Child_2(n)$, and then set $Child_1(n)_{\text{pivot}}$ equal to the centroid of the points owned by $Child_1(n)$ and set $Child_2(n)_{\text{pivot}}$ equal to the centroid of the points owned by $Child_2(n)$. This method has the merit that the cost of splitting $n$ is only linear in the number of points owned by $n$.

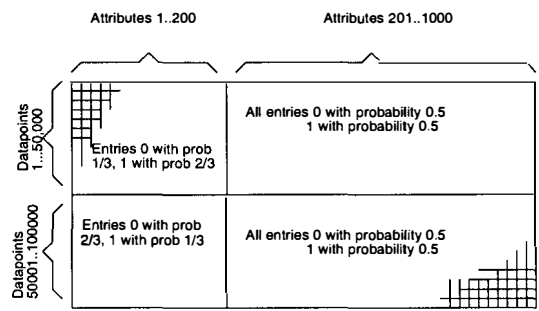

Figure 1: A spreadsheet with 100,000 rows and 1000 columns. In the rightmost 80 percent of the dataset all values are completely random. In the leftmost 20 percent there are more 1's than 0's in the top half and more 0's than 1's in the bottom half. *kd*-trees structure this poorly. Metric trees structure it well.

### 2.1 Why metric trees?

*kd*-trees, upon which earlier accelerations were based, are similar in spirit to decision trees. Each *kd*-tree node has two children, specified by a splitting dimension, $n_{\text{splitdim}}$, and a splitting value, $n_{\text{splitval}}$. A point is owned by child 1 if its $n_{\text{splitdim}}$'th component is less than $n_{\text{splitval}}$, and by child 2 otherwise. Notice *kd*-trees thus need to access vector components directly. *kd*-trees grow top-down. Each node's children are created by splitting on the widest (or highest entropy) dimension.

*kd*-trees are very effective in low dimensions at quickly localizing datapoints: after travelling down a few levels in the tree, all the datapoints in one leaf tend to be much closer to each other than to datapoints not in the same leaf (this loose description can be formalized). That in turn leads to great accelerations when *kd*-trees are used for their traditional purposes (nearest neighbor retrieval and range searching) and also when they are used with cached sufficient statistics. But in high dimensions this property disappears. Consider the dataset in Figure 1. The datapoints come from two classes. In class A, attributes 1 through 200 are independently set to 1 with probability 1/3. In class B, attributes 1 through 200 are independently set to 1 with probability 2/3. In both classes attributes 201 through 1000 are independently set to 1 with probability 1/2. The marginal distribution of each attribute is half 1's and half 0's, so the *kd*-tree does not know which attributes are best to split on. And even if it split on one of the first 200 attributes, its left child would only contain 2/3 of class A and 1/3 of class B (with converse proportions for the right child). The *kd*-tree would thus need to split at least 10 times (meaning thousands of nodes) until at least 99 percent of the datapoints were in a node in which 99 percent of the datapoints were from the same class.

For a metric tree, it is easy to show that the very first split will, with very high probability, put 99 percent of class A into one child and 99 percent of class B into the other. The consequence of this difference is that for a nearest neighbor algorithm, a search will only need to visit half the datapoints in a metric tree, but many more in a $kd$-tree. We similarly expect cached sufficient statistics to benefit from metric trees.

## 3 The Anchors Hierarchy

Before describing the first of our metric-tree-based cached sufficient statistics algorithms, we introduce the *Anchors Hierarchy*, a method of structuring the metric tree that is intended to quickly produce nodes that are more suited to our task than the simple top-down procedure described earlier. As we will see, creating this hierarchy is similar to performing the statistical operation of clustering—one example of the kind of operation we wish to accelerate. This creates a chicken and egg problem: we'd like to use the metric tree to find a good clustering and we'd like to find a good clustering in order to build the metric tree. The solution we give here is a simple algorithm that creates an effective clustering cheaply even in the absence of the tree.

The algorithm maintains a set of *anchors* $A = \{a^1, ..a^k\}$. The $i$'th anchor, $a^i$, has a pivot $a^i{}_{\text{pivot}}$, and an explicit list of the set of points that are closer to $a^i$ than any other anchor:

$$Owned(a^i) = \{x_1^i, x_2^i \ldots x_{n_i}^i\} \qquad (3)$$

where $\forall i, j, p$,

$$x_p^i \in Owned(a^i) \Rightarrow D(x_p^i, a^i{}_{\text{pivot}}) \leq D(x_p^i, a^j{}_{\text{pivot}}) \qquad (4)$$

This list is sorted in decreasing order of distance to $a^i$, and so we can define the *radius* of $a^i$ to be the furthest distance of any of the points owned by $a^i$ to its pivot simply as

$$a^i{}_{\text{radius}} = D(a^i{}_{\text{pivot}}, x_1^i) \qquad (5)$$

At each iteration, a new anchor is added to $A$, and the points it must own (according to the above constraints) are computed efficiently. The new anchor, $a^{\text{new}}$, attempts to steal points from each of the existing anchors. To steal from $a^i$, we iterate through the sorted list of points owned by $a^i$. Each point is tested to decide whether it is closer to $a^i$ or $a^{\text{new}}$. However, if we reach a point in the list at which we inspect $x_p^i$ and we discover that

$$D(x_p^i, a^i) < D(a^{\text{new}}, a^i)/2 \qquad (6)$$

we can deduce that the remainder of the points in $a^i$'s list cannot possibly be stolen because for $k \geq 0$:

$$D(x_{p+k}^i, a^i{}_{\text{pivot}}) \leq D(x_p^i, a^i{}_{\text{pivot}}) \leq$$
$$D(a^{\text{new}}{}_{\text{pivot}}, a^i{}_{\text{pivot}})/2 \leq$$
$$\tfrac{1}{2}D(a^{\text{new}}{}_{\text{pivot}}, x_{p+k}^i) + \tfrac{1}{2}D(x_{p+k}^i, a^i{}_{\text{pivot}})$$

so $D(x_{p+k}^i, a^i{}_{\text{pivot}}) \leq D(a^{\text{new}}{}_{\text{pivot}}, x_{p+k}^i)$. This saving gets very significant when there are many anchors because most of the old anchors discover immediately that none of their points can be stolen by $a^{\text{new}}$.

How is the new anchor $a^{\text{new}}$ chosen? We simply find the current anchor $a^{\text{maxrad}}$ with the largest radius, and choose the pivot of $a^{\text{new}}$ to be the point owned by $a^{\text{maxrad}}$ that is furthest from $a^{\text{maxrad}}$.

This is equivalent to adding each new anchor near an intersection of the Voronoi diagram implied by the current anchor set. Figures 2—6 show an example.

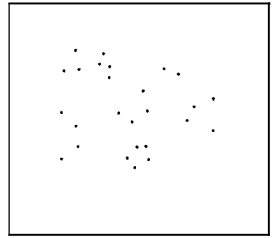

Figure 2: A set of points in 2-d

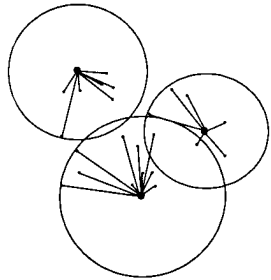

Figure 3: Three anchors. Pivots are big black dots. Owned points shown by rays (the lengths of the rays are explicitly cached). Radiuses shown by circles.

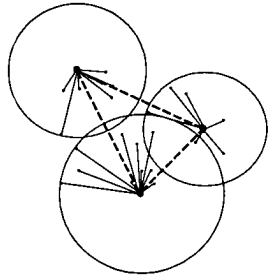

Figure 4: Note that the anchors hierarchy explicitly stores all inter-anchor distances.

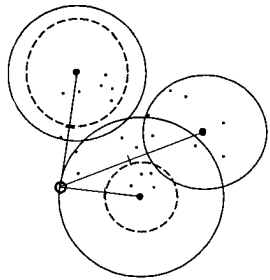

Figure 5: A new anchor is added at the furthest point from any original anchor. Dashed circles show cutoffs: points inside them needn't be checked. None of the furthest circle is checked.

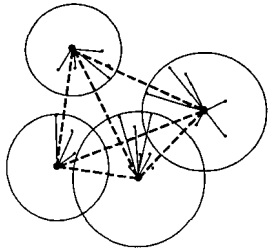

Figure 6: The new configuration with 4 anchors.

To create $k$ anchors in this fashion requires no pre-existing cached statistics or tree. Its cost can be as high as $Rk$ for $R$ datapoints, but in low dimensions, or for non-uniformly-distributed data, it requires an expected $O(R \log k)$ distance comparisons for $k << R$.

### 3.1 Middle-out building of a Cached statistic metric tree

We now describe how we build the metric trees used in this paper. Instead of "top-down" or "bottom-up" (which, for $R$ datapoints is an $O(R^2)$ operation, though can be reduced in cost by approximations) we build it "middle-out". We create an anchors hierarchy containing $\sqrt{R}$ anchors. These anchors are all assigned to be nodes in the tree. Then the most compatible pair of nodes are merged to form a parent node. The compatibility of two nodes is defined to be the radius of the smallest parent node that contains them both completely—the smaller the better.

This proceeds bottom up from the $\sqrt{R}$ anchors until all nodes have been agglomerated into one tree. When this is completed we must deal with the fact that each of the leaf nodes (there are $\sqrt{R}$ of them) contains $\sqrt{R}$ points on average. We will subdivide them further. This is achieved by, for each of the leaves, recursively calling this whole tree building procedure (including this bit), on the set of datapoints in that leaf. The base case of the recursion are nodes containing fewer than $R_{min}$ points. Figures 7—10 give an example.

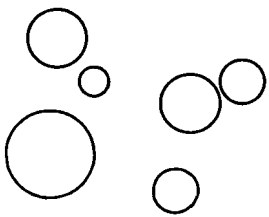

Figure 7: A set of anchors created by the method of the previous section.

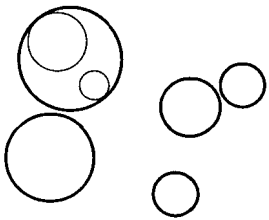

Figure 8: After the first merge step. A new node is created with the two thin-edged circles as children.

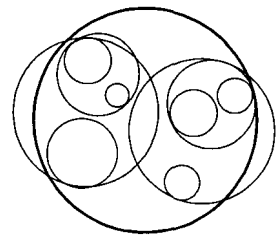

Figure 9: After 5 more merges the root node is created.

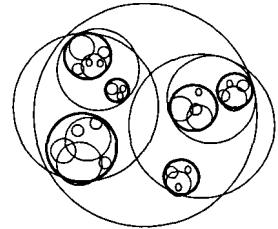

Figure 10: The same procedure is now applied recursively within each of the original leaf nodes.

## 4 Accelerating statistics and learning with metric trees

### 4.1 Accelerating High-dimensional K-means

We use the following cached statistics: each node contains a count of the number of points it owns, and the centroid of all the points it owns[1].

We first describe the naive K-means algorithm for producing a clustering of the points in the input into $K$ clusters. It partitions the data-points into $K$ subsets such that all points in a given subset "belong" to some center. The algorithm keeps track of the centroids of the subsets, and proceeds in iterations. Before the first iteration the centroids are initialized to random values. The algorithm terminates when the centroid locations stay fixed during an iteration. In each iteration, the following is performed:

1. For each point $x$, find the centroid which is closest to $x$. Associate $x$ with this centroid.

2. Re-estimate centroid locations by taking, for each centroid, the center of mass of points associated with it.

The K-means algorithm is known to converge to a local minimum of the distortion measure (that is, average squared distance from points to their class centroids). It is also known to be too slow for very large databases. Much of the related work does not attempt to confront the algorithmic issues directly. Instead, different methods of subsampling and approximation are proposed.

---

[1] In order for the concept of centroids to be meaningful we *do* require the ability to sum and scale datapoints, in addition to the triangle inequality

Instead, we will use the metric trees, along with their statistics, to accelerate K-means with no approximation. This uses a similar approach to (Pelleg & Moore, 1999)'s kd-tree-based acceleration. Each pass of this new efficient K-means recurses over the nodes in the metric tree. The pass begins by a call to the following procedure with $n$ set to be the root node, $C$ set to be the set of centroids, and Cands set to also be the full set of centroids.

*KmeansStep(node n, CentroidSet C, CentroidSet Cands)*

*Invariant: we assume on entry that Cands are those members of $C$ who are provably the only possible owners of the points in node $n$. Formally, we assume that Cands $\subseteq C$, and $\forall x \in n, (argmin_{c \in C} D(x,c)) \in Cands$.*

1. **Reduce *Cands*:** We attempt to prune the set of centroids that could possibly own datapoints from $n$. First, we identify $c^\star \in Cands$, the candidate centroid closest to $n_{\text{pivot}}$. Then all other candidates are judged relative to $c^\star$. For $c \in Cands$, if
$$D(c^\star, n_{\text{pivot}}) + R \leq D(c, n_{\text{pivot}}) - R$$
then delete $c$ from the set of candidates, where $R = n_{\text{radius}}$ is the radius of node $n$. This cutoff rule will be discussed shortly.

2. **Update statistics about new centroids:** The purpose of the K-means pass is to generate the centers of mass of the points owned by each centroid. Later, these will be used to update all the centroid locations. In this step of the algorithm we accumulate the contribution to these centers of mass that are due to the datapoints in $n$. Depending on circumstances we do one of three things:

   - If there is only one candidate in *Cands*, we need not look at $n$'s children. Simply use the information cached in the node $n$ to award all the mass to this one candidate.
   - Else, if $n$ is a leaf, iterate through all its datapoints, awarding each to its closest centroid in *Cands*. Note that even if the top-level call had $K = 1000$ centroids, we can hope that at the levels of leaves there might be far fewer centroids remaining in *Cands*, and so will see accelerations over conventional K-means even if the search reaches the leaves.
   - Else recurse. Call **KmeansStep** on each of the child nodes in turn.

The candidate removal test in Step 1 is easy to understand. If that test is satisfied then for any point $x$ in the node

$$D(x, c^\star) \leq D(x, n_{\text{pivot}}) + D(n_{\text{pivot}}, c^\star) \leq$$
$$R + D(n_{\text{pivot}}, c^\star) \leq$$
$$D(c, n_{\text{pivot}}) - R \leq D(c, x) + D(x, n_{\text{pivot}}) - R \leq$$
$$D(c, x) + R - R = D(c, x)$$

and so $c$ cannot own any $x$ in $n$.

### 4.2 Example 2: Accelerating Non-parametric Anomaly Detection

As an example of a non-parametric statistics operation that can be accelerated, consider a test for anomalous datapoints that proceeds by identifying points in low density regions. One such test consists of labeling a point as anomalous if the number of neighboring points within some radius is less than some threshold.

This operation requires counts within the nodes as the only cached statistic. Given a query point $x$, we again recurse over the tree in a depth first manner, trying the child closer to $x$ before the further child. We maintain a count of the number of points discovered so far within the range, and another count—an upper bound on the number of points that could possibly be within range. At each point in the search we have an impressive number of possibilities for pruning the search:

1. If the current node is entirely contained within the query radius, we simply add $n_{\text{count}}$ to the count.

2. If the current node is entirely outside the query radius, we simply ignore this node, decrementing the upper bound on the number of points by $n_{\text{count}}$.

3. If the count of the number of points found ever exceeds the threshold, we simply return FALSE—the query is not an anomaly.

4. If the upper bound on the count ever drops below threshold then return TRUE—the query is an anomaly.

### 4.3 Example 3: Grouping Attributes

The third example is designed to illustrate an important use of high dimensional methods. Consider a dataset represented as a matrix in which rows correspond to datapoints and columns correspond to attributes. Occasionally, we may be more interested in groupings among attributes than among datapoints. If so, we can transpose the dataset and build the metric tree on attributes instead of datapoints.

For example, suppose we are interested in finding pairs of attributes that are highly positively correlated. Remember that the correlation coefficient between two attributes $x$ and $y$ is

$$\rho(x,y) = \sum_i (x_i - \bar{x})(y_i - \bar{y})/(\sigma_x \sigma_y) \qquad (7)$$

where $x_i$ is the value of attribute $x$ in the $i$th record. If we subtract the attribute means and divide by their standard deviations we may define normalized values $x_i^\star = (x_i - \bar{x})/\sigma_x$ and $y_i^\star = (y_i - \bar{y})/\sigma_y$. Then

$$\rho(x,y) = \sum_i x_i^\star y_i^\star = 1 - D^2(x^\star, y^\star)/2 \qquad (8)$$

using Euclidean distance. Thus, for example, finding all pairs of attributes correlated above a certain level $\rho$ corresponds to finding pairs with distances below $\sqrt{2-2\rho}$.

There is no space to describe the details of the algorithm using metric trees to collect all these close pairs. It is a special case of a general set of "all-pairs" algorithms described (albeit on conventional $kd$-trees) by (Gray & Moore, 2000) and related to the celebrated Barnes-Hut method (Barnes & Hut, 1986) for efficient $R$-body simulation.

## 5 Empirical Results

The following results used the datasets in Table 1.

Each dataset had K-means applied to it, the non-parametric anomaly detector, and all-pairs. For the latter two operations suitable threshold parameters were chosen so that the results were "interesting" (e.g. in the case of anomalies, so that about 10 percent of all datapoints were regarded as anomalous). This is important: all the algorithms do very well (far greater accelerations) in extreme cases because of greater pruning possibilities, and so our picking of "interesting" cases was meant to tax the acceleration algorithms as much as possible. For K-means, we ran three experiments for each dataset in which $K$ (the number of centroids) was varied between 3, 20 and 100.

The results in Table 2 show three numbers for each experiment. First, the number of distance comparisons needed by a regular (i.e. treeless) implementation of the algorithm. Second, the number of distance comparisons needed by the accelerated method using the cached-statistics-supplemented metric tree. And third, in bold, the speedup (the first number divided by the second). For the artificial datasets, we restricted K-means experiments to those in which $K$, the number of clusters being searched for, matched the actual number of clusters in the generated dataset.

The speedups in the two dimensional datasets are all excellent, but of course in two dimensions regular statistics-caching $kd$-trees would have given equal or better results. The higher dimensional cell and covtype datasets are more interesting—the statistics caching metric trees give a substantial speedup whereas in other results we have established that $kd$-trees do not. Notice that in several results, K=20 gave a worse speedup than K=3 or K=100. Whether this is significant remains to be investigated.

The Reuters dataset gave poor results. This only has 10,000 records and appears to have little intrinsic structure. Would more records have produced better results? Extra data was not immediately available at the time of writing and so, to test this theory, we resorted to *reducing* the amount of data. The **Reuters50** dataset has only half the documents, and its speedups are indeed worse. This gives some credence to the argument that if the amount of data was increased the anti-speedup of the metric trees would be reduced, and eventually might become profitable.

On anomaly detection and all-pairs, all methods performed well, even on Reuters. There is wild variation in the speedup. This is an artifact of the choice of thresholds for each algorithm. If too large or too small pruning becomes trivial and the speedup is enormous.

Table 3 investigates whether the "anchors" approach to building the tree has any advantage over the simpler top-down approach. For the four datasets we tested, using K-means as the test, the speedups of using the anchors-constructed tree over the top-down-constructed tree were modest (ranging from 20 percent to 180 percent) but consistently positive. Similar results comparing the two tree growth methods for all-pairs and anomalies give speedups of 2-fold to 6-fold.

As an aside, Table 4 examines the quality of the clustering created by the anchors algorithm. In all experiments so far, K-means was seeded with random centroids. What if the anchors algorithm was used to generate the starting centroids? Table 4 shows, in the middle four columns, the distortion (sum squared distance from datapoints to their nearest centroids) respectively for randomly-chosen centroids, centroids chosen by anchors, centroids started randomly followed by 50 iterations of K-means and centroids started with anchors followed by 50 iterations of K-means. Both before and after K-means, anchors show a substantial advantage except for the Reuters dataset.

| Dataset | No. of data-points, $R$ | No. of dimensions, $M$ | Description |
|---|---|---|---|
| squiggles | 80000 | 2 | Two dimensional data generated from blurred one-dimensional manifolds |
| voronoi | 80000 | 2 | Two dimensional data with noisy filaments |
| cell | 39972 | 38 | Many visual features of cells observed during high throughput screening |
| covtype | 150000 | 54 | Forest cover types (from (Bay, 1999)) |
| reuters100 | 10077 | 4732 | Bag-of-words representations of Reuters news articles (from (Bay, 1999)) |
| gen$M$-k$i$ | 100000 | $M$ | Artificially generated sparse data in $M$ dimensions, generated from a mixture of $i$ components |

Table 1: Datasets used in the empirical results.

| | | k=3 | k=20 | k=100 | All Pairs | Anomalies |
|---|---|---|---|---|---|---|
| squiggles | regular | 4.08e+07 | 2.72e+08 | 1.36e+09 | 3.19e+09 | 3.20e+09 |
|  | fast | 8.25e+05 | 4.03e+06 | 8.55e+06 | 2.17e+06 | 3.38e+06 |
|  | speedup | **49.4** | **67.4** | **158** | **1474** | **946** |
| voronoi | regular | 4.08e+07 | 2.72e+08 | 1.36e+09 | 3.20e+09 | 3.20e+09 |
|  | fast | 9.25e+05 | 6.24e+06 | 1.54e+07 | 8.95e+05 | 8.12e+06 |
|  | speedup | **44.1** | **43.6** | **88.4** | **3574** | **394** |
| cell | regular | 1.32e+07 | 8.79e+07 | 4.40e+08 | 7.99e+08 | 7.99e+08 |
|  | fast | 1.17e+06 | 1.25e+07 | 3.92e+07 | 8.14e+06 | 2.44e+07 |
|  | speedup | **11.3** | **7.0** | **11.2** | **98.1** | **32.7** |
| covtype | regular | 4.95e+07 | 3.30e+08 | 1.65e+09 | 1.12e+10 | 1.12e+10 |
|  | fast | 1.99e+06 | 2.91e+07 | 8.69e+07 | 1.44e+08 | 4.86e+07 |
|  | speedup | **24.8** | **11.3** | **19.0** | **78.2** | **231** |
| reuters50 | regular | 1.31e+06 | 8.74e+06 | 4.37e+07 | 1.27e+07 | 1.27e+07 |
|  | fast | 2.05e+06 | 1.28e+07 | 6.66e+07 | 1.65e+07 | 3.70e+07 |
|  | speedup | **0.6** | **0.7** | **0.7** | **0.8** | **0.3** |
| reuters100 | regular | 2.62e+06 | 1.75e+07 | 8.74e+07 | 5.08e+07 | 5.08e+07 |
|  | fast | 3.06e+06 | 2.04e+07 | 1.02e+08 | 2.03e+07 | 3971 |
|  | speedup | **0.9** | **0.9** | **0.9** | **2.5** | **1.28e+04** |
| gen100-k3 | regular | 3.30e+07 | - | - | 5.00e+09 | 1.00e+10 |
|  | fast | 2.53e+06 | - | - | 231 | 1.70e+07 |
|  | speedup | **13.0** | - | - | **2.16e+07** | **588** |
| gen100-k20 | regular | - | 2.20e+08 | - | 5.00e+09 | 1.00e+10 |
|  | fast | - | 4.59e+07 | - | 57.0 | 5.00e+06 |
|  | speedup | - | **4.8** | - | **8.77e+07** | **2000** |
| gen100-k100 | regular | - | - | 1.08e+09 | 5.00e+09 | 1.00e+10 |
|  | fast | - | - | 3.02e+08 | 1464 | 3.10e+06 |
|  | speedup | - | - | **3.6** | **3.42e+06** | **3220** |
| gen1000-k3 | regular | 3.30e+07 | - | - | 5.00e+09 | 1.00e+10 |
|  | fast | 3.97e+06 | - | - | 7.16e+08 | 3.40e+07 |
|  | speedup | **8.3** | - | - | **7.0** | **294** |
| gen1000-k20 | regular | - | 2.13e+08 | - | 5.00e+09 | 1.00e+10 |
|  | fast | - | 2.54e+07 | - | 1.57e+08 | 5.70e+06 |
|  | speedup | - | **8.4** | - | **31.9** | **1754** |
| gen1000-k100 | regular | - | - | 1.07e+09 | 5.00e+09 | 1.00e+10 |
|  | fast | - | - | 3.30e+08 | 3.39e+07 | 1.40e+06 |
|  | speedup | - | - | **3.2** | **147** | **7143** |
| gen10000-k3 | regular | 3.30e+07 | - | - | 5.00e+09 | 1.00e+10 |
|  | fast | 1650 | - | - | 7.13e+08 | 2.90e+07 |
|  | speedup | **2.00e+04** | - | - | **7.0** | **344** |
| gen10000-k20 | regular | - | 2.20e+08 | - | 5.00e+09 | 1.00e+10 |
|  | fast | - | 7.29e+07 | - | 1.57e+08 | 5.90e+06 |
|  | speedup | - | **3.0** | - | **31.9** | **1694** |
| gen10000-k100 | regular | - | - | 1.10e+09 | 5.00e+09 | 1.00e+10 |
|  | fast | - | - | 4.47e+08 | 3.41e+07 | 4.90e+06 |
|  | speedup | - | - | **2.5** | **146** | **2040** |

Table 2: Results on various datasets. Non-bold numbers show the number of distance computations needed in each experiment. Bold-face numbers show the speedup—how many times were the statistics-caching metric trees faster than conventional implementations in terms of numbers of distance computations? The columns labeled $k=3$, $k=20$ and $k=100$ are for K-means.

|  |  | Random Start | Anchors Start | Random End | Anchors End | Start Benefit | End Benefit |
|---|---|---|---|---|---|---|---|
| cell | k=100 | 2.40174e+13 | 8.00569e+12 | 3.81462e+12 | 3.21197e+12 | 3.00 | 1.19 |
|  | k=20 | 1.45972e+14 | 3.44561e+13 | 1.81287e+13 | 1.16092e+13 | 4.24 | 1.56 |
|  | k=3 | 1.84008e+14 | 8.89971e+13 | 2.16672e+14 | 1.01674e+14 | 2.07 | 2.13 |
| covtype | k=100 | 6.59165e+09 | 4.08093e+09 | 4.70664e+09 | 4.04747e+09 | 1.4005 | 1.00827 |
|  | k=20 | 3.06986e+10 | 1.09031e+10 | 1.29359e+10 | 1.04698e+10 | 2.37313 | 1.04139 |
|  | k=3 | 1.48909e+11 | 6.09021e+10 | 7.04157e+10 | 6.09162e+10 | 2.11471 | 0.999769 |
| reuters100 | k=100 | 11431.5 | 6455.6 | 6531.8 | 6428.09 | 1.75013 | 1.00428 |
|  | k=20 | 12513.5 | 6672.24 | 6773.7 | 6661.73 | 1.84737 | 1.00158 |
|  | k=3 | 13401.1 | 6890.97 | 6950.35 | 6880.76 | 1.92812 | 1.00148 |
| squiggles | k=100 | 180.369 | 75.0007 | 64.452 | 54.9265 | 2.7985 | 1.36547 |
|  | k=20 | 1269.4 | 589.974 | 511.912 | 466.93 | 2.47972 | 1.26352 |
|  | k=3 | 13048.3 | 4821.91 | 4252.64 | 4109.01 | 3.06828 | 1.1735 |

Table 4: The numbers in the four central columns show the distortion measure for a variety of experiments. "Random Start" are the distortion measures for randomly chosen centroids. "Anchors Start" are the distortion measures obtained when using the anchors hierarchy to generate initial centroids. The next two columns show the resulting distortion of random-start and anchors-start respectively after 50 iterations of K-means. The final two columns summarize the relative merits of anchors versus random: by what factor is anchor's distortion better than random's? "Start Benefit" shows the factor for the initial centroids. "End benefit" shows the factor for the final centroids.

| Dataset | k=3 | k=20 | k=100 |
|---|---|---|---|
| cell | 1.3 | 1.2 | 1.2 |
| covtype | 1.3 | 1.3 | 1.3 |
| squiggles | 1.6 | 1.5 | 1.6 |
| gen10000-k20 | 2.8 | 2.7 | 2.7 |

Table 3: The factor by which using anchors to build the metric tree improves over using top-down building in terms of number of distance calculations needed.

## 6   Accelerating other learning algorithms

The three cached-statistic metric tree algorithms introduced in this paper were primarily intended as examples. A very wide range of both parametric and non-parametric methods are amenable to acceleration in a similar way. As further examples, and on our short-list for future work are:

- Dependency trees (Meila, 1999), by modifying the all-pairs method above to the case of Kruskal's algorithm for minimum spanning tree in Euclidean space (equivalently maximum spanning tree in correlation space). Additionally, although Mutual Information does not obey the triangle inequality, it can be bounded above (Meila, 2000) and below by Euclidean distance and so it should also be possible to apply this method to entropy based tree building for high-dimensional discrete data.

- Mixtures of spherical, axis aligned or general Gaussians. These are all modifications of the K-means algorithm above and the *mrkd*-tree-based acceleration of mixtures of Gaussians described in (Moore, 1999).

- Mixtures of multinomials for high dimensional discrete data.

- There are numerous other non-parametric statistics, including $n$-point correlation functions used in astrophysics that we will apply these techniques to. In addition, Gaussian processes, certain neural net architectures, and case-based reasoning systems may be accelerated.

## 7   Discussion

The purpose of this paper has been to describe, discuss and empirically evaluate the use of metric trees to help statistical and learning algorithms scale up to datasets with large numbers of records and dimensions. In so doing, we have introduced the anchors hierarchy for building a promising first cut at a set of tree nodes efficiently before the tree has been created. We have given three examples of cached-sufficient-statistics-based algorithms built on top of these structures.

If there *is* no underlying structure in the data (e.g. if it is uniformly distributed) there will be little or no acceleration in high dimensions no matter what we do. This gloomy view, supported by recent theoretical work in computational geometry (Indyk, Amir, Efrat, & Samet, 1999), means that we can only accelerate datasets that have interesting internal structure. Resorting to empirical results with real datasets, however, there is room for some cautious optimism for real-world use.